\documentclass[letterpaper, 10 pt, conference]{ieeeconf}  

\IEEEoverridecommandlockouts                              

\usepackage{graphics} 
\usepackage{amsmath} 
\usepackage{amssymb}  

\usepackage[noadjust]{cite}

\usepackage{moreverb,url}

\usepackage{subcaption}
\usepackage{makecell}
\usepackage{adjustbox}
\usepackage{pifont}
\usepackage{multirow}
\usepackage[table,xcdraw]{xcolor}
\usepackage[font=small,labelfont=bf]{caption}
\usepackage{hyperref}

\usepackage{tikz}
\newcommand\copyrightnotice[1]{
    \begin{tikzpicture}[remember picture,overlay]
    \node[anchor=north,xshift=-65,yshift=-12pt] at (current page.north) {\parbox{\dimexpr0.75\textwidth-\fboxsep-\fboxrule\relax}{\scriptsize #1}};
    \end{tikzpicture}
}

\newcommand{\hide}[1]{}

\title{\LARGE \bf
Context-aware collaborative pushing of heavy objects \\using skeleton-based intention prediction
}

\author{Gokhan Solak$^{1}$, Gustavo J. G. Lahr$^{2}$, Idil Ozdamar$^{1}$, and Arash Ajoudani$^{1}$
\thanks{This work was supported by the European Union Horizon Project
TORNADO (GA 101189557).}
\thanks{$^{1}$Human-Robot Interfaces and Interaction Lab, Istituto Italiano di Tecnologia, Genoa, Italy.
        {\tt\small gokhan.solak@iit.it}}%
\thanks{$^{2}$Instituto Israelita de Ensino e Pesquisa, Hospital Israelita Albert Einstein, São Paulo, Brazil.}%
}

\begin{document}

\maketitle
\thispagestyle{empty}
\pagestyle{empty}

\begin{abstract}

In physical human-robot interaction, force feedback has been the most common sensing modality to convey the human intention to the robot.
It is widely used in admittance control to allow the human to direct the robot. 
However, it cannot be used in scenarios where direct force feedback is not available since manipulated objects are not always equipped with a force sensor. In this work, we study one such scenario: the collaborative pushing and pulling of heavy objects on frictional surfaces, a prevalent task in industrial settings. 
When humans do it, they communicate through verbal and non-verbal cues, where body poses, and movements often convey more than words.
We propose a novel context-aware approach using Directed Graph Neural Networks to analyze spatio-temporal human posture data to predict human motion intention for non-verbal collaborative physical manipulation. Our experiments demonstrate that robot assistance significantly reduces human effort and improves task efficiency. The results indicate that incorporating posture-based context recognition, either together with or as an alternative to force sensing, enhances robot decision-making and control efficiency.
\end{abstract}

\copyrightnotice{\copyright 2025 IEEE.  Personal use of this material is permitted.  Permission from IEEE must be obtained for all other uses, in any current or future media, including reprinting/republishing this material for advertising or promotional purposes, creating new collective works, for resale or redistribution to servers or lists, or reuse of any copyrighted component of this work in other works.}

\section{INTRODUCTION}


Predicting human intention is critical for integrating robots into human environments. Humans seamlessly communicate their intentions verbally or non-verbally (e.g., vision, haptics), exchanging information and intentions through experiences. Especially in collaborative tasks, humans communicate naturally through body poses and movements rather than explicit communication~\cite{Stergiou2019human-human-interaction}. For example, when two people carry a heavy object together, they rely on haptic feedback, body language, and task understanding rather than continuous verbal instructions. The information combined among agents, environments, and tasks defines a context that enables a shared goal to be achieved~\cite{Liu2021coexistence}.

\begin{figure}[t]
    \centering
    \includegraphics[width=0.91\linewidth]{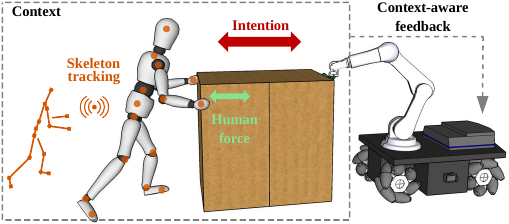}
    \caption{The task is split into three possible movements: when the human intends to push, pull, or stay idle. The collaborative robot predicts the intended motion from the human skeleton tracking data to act appropriately in the context. 
    }
    \label{fig:task-concept}
    \vspace{-0.5cm}
\end{figure}

Using direct force feedback is a natural and effective way of achieving non-verbal communication. 
Accordingly, admittance control has been the predominant approach in physical human-robot interaction (pHRI) \cite{ajoudani2018progress}. 
However, direct force feedback is not always available, such as when the manipulated object is deformable or when human force interacts with external forces. In those cases, the robot must gain context awareness through other modalities.
We study this problem through the novel collaborative task of pushing and pulling heavy objects on a frictional surface, illustrated in Fig.~\ref{fig:task-concept}.
It is a task with high demand in industry~\cite{humanBodyMechanics2017} and preferred in situations where objects are bulky or heavy~\cite{Rajendran2021ergonomicsLifting}. 

Moving heavy objects on a frictional surface is a common task that human pairs tackle collaboratively by holding the object on opposing sides and applying force in the same direction by pushing or pulling. Synchronizing the force timing and direction is vital to the task's success. Unlike the classic collaborative carrying problem~\cite{Agravante2019collabCarry}, the forces cannot be directly transmitted to the partner since they are subjected to the friction force, as explained in Sec.~\ref{sec:pull-push}. Also a human-side sensor is not feasible as it requires either equipping each object with a force sensor or measuring human wrist force at the world frame.
Common haptic gloves have sensors on their fingertips that measure the grasping force, rather than the wrist force, as we need in this problem.

The efficiency and success of this task heavily depend on the robot's ability to understand and predict the human operator's intentions. 
Human pose, as a non-verbal communication cue, presents a rich source of information that can be leveraged for intention prediction in pushing and pulling as humans adapt their body pose accordingly~\cite{Ajoudani2017choosingPoses}. 
Raw skeleton data cannot be directly used in the decision-making process.
To address this, we employ a context-aware approach by capturing human temporal data and training a Directed Graph Neural Network (DGNN)~\cite{Shi2019dgnn} to infer task timing and human intention. 


The primary objective of this paper is to bridge the identified gap with a novel approach that integrates context-aware human intention detection using pose analysis for a pHRI task within dynamic environments. Specifically, we present it on the novel task of collaborative pushing large objects on frictional surfaces. Our real-world robotic experiments show that our method facilitates natural and smooth HRI during the task, which involves pushing, pulling, and the transitional motions in between. The context-aware robot assistance significantly reduces human effort without any training. 
Our study adds valuable insights into human-robot collaboration in dynamic tasks, which require agents to better understand their nonverbal cues and combine their efforts.


\section{Related Works}\label{sec:related-works}


Interpreting human intentions requires more than merging sensor data~\cite{Losey2018pHRIsurvey}. There is a need to develop techniques to understand the task context~\cite{Wang2024vision, Liu2021coexistence}. Multi-modal context-awareness in pHRI is not a well-explored topic in the robotics literature. The past works mainly focused on the coexistence of robots and humans such as collision avoidance~\cite{Liu2021coexistence, Nikolakis2018coexistence}, hand gesture recognition~\cite{Wang2024vision} and investigating social-cognitive interactions~\cite{Quintas2019cognitive}. 
Approaches that learn physical interaction context depend mainly on direct force feedback \cite{drolet2023learning} or use other modalities such as gaze tracking to augment the primarily force-based interaction \cite{Wong2023vision-and}.
Unlike the existing works, we focus on indirect pHRI where force feedback is not sufficient to solve the problem. 
We explore the collaborative pushing and pulling task that necessitates context-awareness while being physical. 

The task of moving objects by sliding them has been studied using humanoids and humanoid-like~\cite{Vaz2020humanoid1, Polverini2020humanoid2, Murooka2015humanoid3, Saeedvand2021humanoid4}, wheeled robots~\cite{Kolhe2010wheeled1,  Ozdamar2024MobilePushing,  Bertoncelli2020wheeled3}, and by robotic manipulators~\cite{Zhou2016manipulator1, Stuber2020survey, Wang2023manipulator2}. However, collaborative scenarios remain unexplored in which a human and robot must slide an object due to inability or preference (e.g., greater energy efficiency). In industries like automotive, 10\% of tasks involve pushing and pulling, with 40\% of those tasks requiring movement of objects over 200 kg~\cite{humanBodyMechanics2017, Rajendran2021ergonomicsLifting}, indicating a need for robotic systems to assist humans in collaboratively pushing and pulling objects.

Due to its complexity and wide range of applications, collaborative object carrying has been a subject of interest for several studies among HRI~\cite{Agravante2019collabCarry}. The benchmark involves a human and a robot holding a rigid object off the ground to move the object toward a desired position. This scenario can be varied in several ways, including overcoming different obstacles~\cite{Ng2024PolicyObstacle, Sirintuna2024Obstacle-awareCollabTransportation}, moving objects in a virtual environment through haptic interfaces~\cite{Madan2015hapticsCarry}, multiple agents~\cite{Sirintuna2023MultiAgent}, and non-rigid objects co-carrying~\cite{Sirintuna2024DeformationCarry}.

The complexity of detecting human intention increases significantly when an object is placed on a surface compared to when it is carried. Consider the benchmark scenario where the intention is identified through the exchange of forces measured by a force-torque sensor at the robot's wrist~\cite{Ng2024PolicyObstacle, Agravante2019collabCarry}. However, detecting the human's intention of movement using only a force-torque sensor when sliding is impractical. For instance, when the object is moving, friction forces, inertia, and the combined forces exerted by both the robot and the human result in a net force that the sensor measures, making it impossible to decouple the human force. When the velocity is null, the sensor at the robot's wrist will not measure any forces until the object starts to move and overcomes static friction~\cite {Bona2005frictionOverview}. 

Current studies predominantly address scenarios where robots and humans lift and carry objects, using force-torque sensors to detect intentions or robots autonomously execute the task. However, these approaches face significant limitations when applied to sliding tasks due to the complexity of accurately interpreting the combined forces. Our research aims to solve this problem by leveraging human body pose tracking to achieve context-aware human-robot collaboration.

\begin{figure}[t]
    \centering
    \includegraphics[width=.99\linewidth]{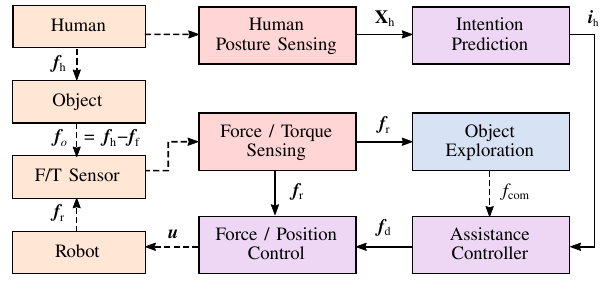}
    \caption{The information flow in our system. The independent bodies of the human, object, and robot interact by direct forces $\boldsymbol{f}_{h}, \boldsymbol{f}_{o}$ and $\boldsymbol{f}_{r}$. Human intention is predicted using human posture sensing, and object friction information is discovered using the robot-side F/T sensor. Intention vector $\boldsymbol{i}_h$ and compensation force ${f}_{com}$ are used by the assistance controller to generate the desired robot forces $\boldsymbol{f}_d$ that are aligned with the expected human forces.}
    \label{fig:control-flow}
    \vspace{-0.4cm}
\end{figure}

\section{Collaborative pull-push}\label{sec:pull-push}

The collaborative pushing problem involves interaction between 3 bodies: the human, the object, and the robot. Although the object is passive, it reacts to the other entities due to frictional force and inertia. 
We illustrate the flow of forces between these bodies in Fig.~\ref{fig:control-flow}. From the robot's perspective, it is impossible to distinguish the human force and friction force using its force-torque (F/T) sensor. 
The sensor measures the compression force between the object and the robot ($\boldsymbol{f}_{r}$) which is the net value of the forces applied by the human to the object ($\boldsymbol{f}_{h}$), the friction forces between object and surface ($\boldsymbol{f}_{f}$), in quasi-static case:

\begin{equation}\label{eq:force-robot-object}    
\boldsymbol{f}_{r}=-\boldsymbol{f}_{h}+\boldsymbol{f}_{f}.
\end{equation}

Thus, $\boldsymbol{f}_{h}$ cannot be decoupled solely from the sensor's measurements. 
Consequently, an admittance controller is unfeasible for the collaborative pushing task, as it requires human force information to render the desired admittance to the human operator.

Since it is impractical to attach sensors to every moveable object or measuring the human wrist force directly, we answer this problem by tracking the human posture. This approach also enables choosing the most convenient grasp location for moving the object rather than being restricted by where sensors are attached. We leverage this information to predict the human intention (Sec.\ref{sec:intention-prediction}) and set the context of the assistive controller (Sec.~\ref{sec:assistive-controller}).  

We track the human posture in form of skeleton data, collected using wearable inertial measurement units (IMU)-based Xsens sensors, without any camera.
Other human pose measurement techniques, such as RGB cameras, might also be used for this purpose, bringing their own advantages and disadvantages \cite{sun2022human}.

In summary, our system uses two data sources, human posture and robot-side force, to achieve context-aware collaborative pushing as illustrated in Fig.~\ref{fig:control-flow}. 
We use the robot's F/T sensor for a brief object exploration without the human. Temporally removing the human factor $\boldsymbol{f}_{h}$ allows us to identify $f_{com}$, the absolute force value that compensates for the static friction for a given object. 
The exploration is done once per object. Later, the F/T sensor is used as feedback for the force controller during the manipulation (Sec.~\ref{sec:force-control}). 
The assistance controller uses both human and object information to set the force targets $\boldsymbol{f}_d$, which is converted to low-level control signal $\boldsymbol{u}$ by the force controller.

In the following sections, we first analyze the task and then introduce the components of our system.

\subsection{Task modeling}

The task of repositioning heavy objects is depicted in Fig. \ref{fig:task-concept}. We assume an object that is rigid and too heavy to be lifted, demanding to be moved by exerting forces in the direction of the desired movement. The movement is executed along 1-DoF for $d$ meters, either pushing ($-x$) or pulling ($+x$). In both cases, the robot is supposed to identify the human's intention and provide help for the task. Although the proof of concept here is done in 1-DoF, the formulation of our method is presented in matrix form to depict the general feasibility of extending it to planar motion.


The friction forces $\boldsymbol{f}_{f}$ are modeled mostly as Coulomb's model \cite{Bona2005frictionOverview}. When the object has null velocity, the friction force has a value between the net external forces applied to it (in this case, the human and robot's forces) and the peak of the static friction $\mu_s \boldsymbol{f}_n$, where $\mu_s$ is the static friction coefficient and $\boldsymbol{f}_n$ is the object's weight. 
Looking through the task point of view, one may notice that for the robot to start giving assistance through a purely reactivate controller, such as admittance or force control, the human should first apply enough force to break static friction and start moving the object, until the robot can measure any force $\boldsymbol{f}_{r}{>}0$. If the object is heavy, breaking static friction may be difficult and tiring for humans. This poses an additional reason to use a human motion intention prediction approach, in addition to the information decoupling issue mentioned earlier.
\subsection{Intention prediction}\label{sec:intention-prediction}

\begin{figure}[t]
    \centering
    \includegraphics[width=.98\linewidth]{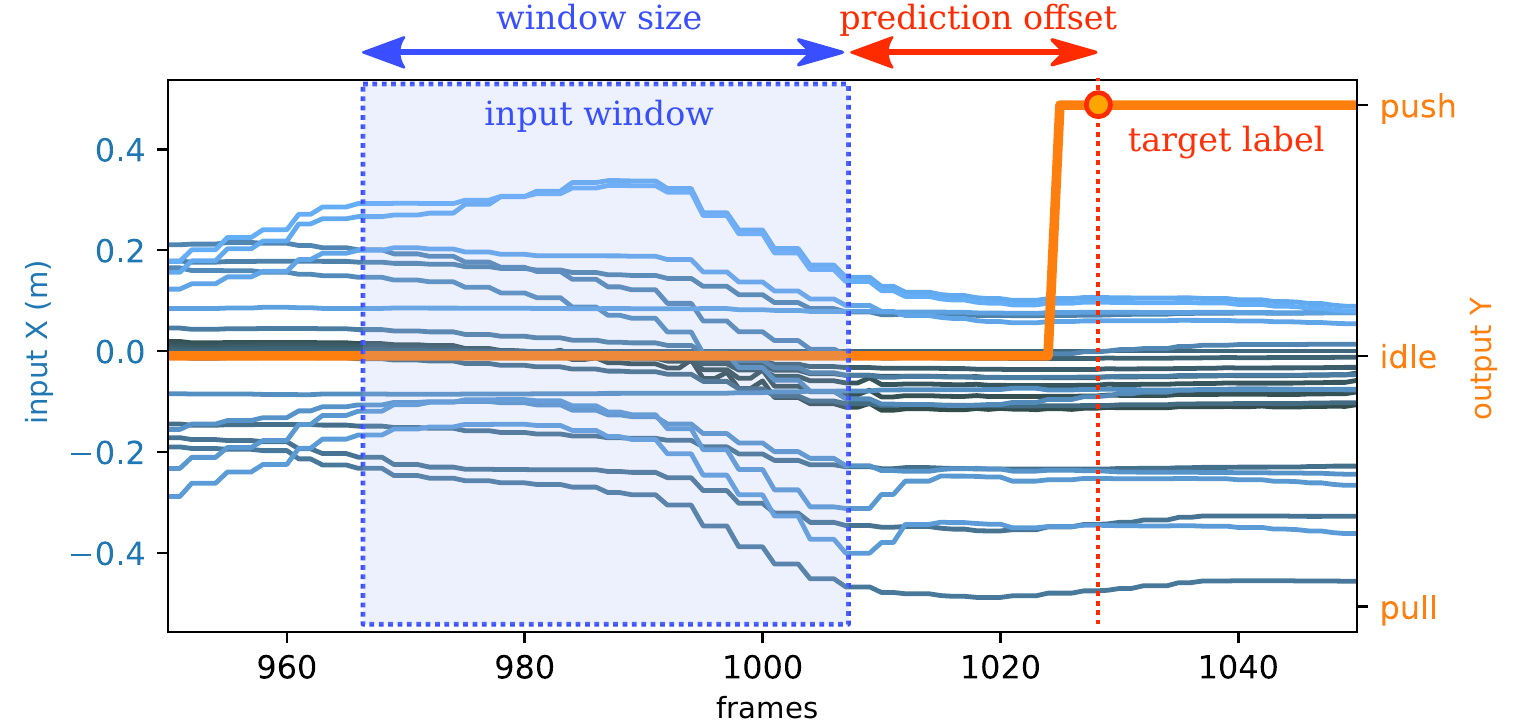}
    \caption{We formulate the intention prediction problem as computing the output class (orange: \textit{push/idle/pull}) given a time window of input features (blue: 3D joint and bone data). 
    In this work, we set the window size and prediction offset as $0.5$ \textit{s} and $0.25$ \textit{s}.}\label{fig:prediction-task}
    \vspace{-0.2cm}
\end{figure}

\begin{figure}[t]
    \centering
    \includegraphics[width=.96\linewidth]{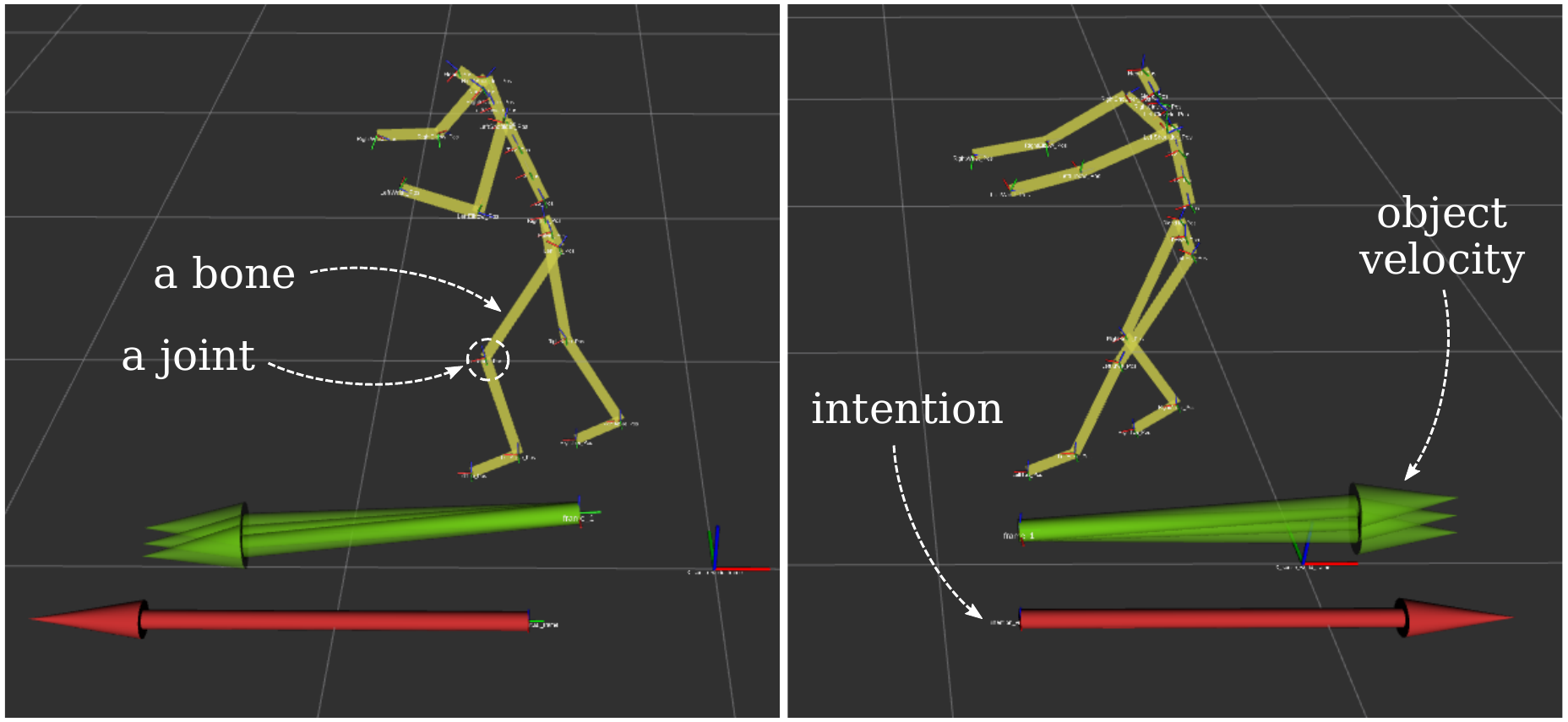}
    \caption{Visualisation of the push (left) and pull (right) actions during runtime. The measured object velocity (green) and the intention prediction output (red) are also shown as arrows. The figure also shows examples of bone and joint data.}\label{fig:rviz-intention}
    \vspace{-0.5cm}
\end{figure}

We illustrate the real-time intention prediction task in Fig.~\ref{fig:prediction-task}: Given a time window of input data $\boldsymbol{X}$, compute the human intention label $\boldsymbol{Y}$ after a given time offset. This can be modeled as a supervised learning problem (classification or regression). 
In this work, we use the skeleton data $\boldsymbol{X}_h$ as the input because it preserves the meaningful structure of the human posture information. $\boldsymbol{X}_h$ includes the joint (3-D position) and the bone data (difference between the connected joint positions) of the human as shown in Fig.~\ref{fig:rviz-intention}.
The output is the intention vector $\boldsymbol{i}_h$ that indicates the desired motion for each dimension (\{\textit{pull}, \textit{idle}, \textit{push}\} or $\{-1, 0, +1\}$). In this work, we study the one-dimensional case as a proof of concept, thus, only the $x$-dimension is predicted, and others are assumed to be zero.

We adopt the DGNN method~\cite{Shi2019dgnn}, a state-of-the art action recognition method. It retains the structure of the skeleton data in the form of a directed graph and leverages the dependencies between joints and bones.
The network updates attributes of vertices and edges in multiple layers, extracting local information in lower layers and more global semantic information in higher layers. Temporal dynamics are modeled using convolutions along the temporal dimension, which decouples spatial and temporal modeling, allowing efficient and effective spatiotemporal information extraction. 

This method was originally used for classifying action sequences where the whole sequence had a single label \cite{Shi2019dgnn}. In our case, the action label changes at each time frame, thus we re-model the problem as shown in Fig.~\ref{fig:prediction-task} where the prediction is done online in real-time. 

We used a shallower neural network than \cite{Shi2019dgnn} as our problem contains 3 classes compared to the 60 classes. For the best fit, we used a network of 3 graph temporal networks (GTN) of 32, 64, and 64 channels, respectively. We also applied dropout ($p=0.3$) after the 3$^{rd}$ GTN and finished with 2 fully connected layers. The network has 3 outputs for 3 classes $y_j, j=1,2,3$. The class with the highest value is selected: $i_h = \arg\!\min_j y_j$.
We also added a preprocessing step to transform all joint data to have the pelvis as the reference frame. This allowed our data to be independent of the starting position. We discuss the training results later in Sec.~\ref{sec:results} after introducing our dataset in Sec.~\ref{sec:dataset}, however, our design choices for preprocessing and architecture are influenced by the training performance.


\begin{figure}[t]
    \centering
    \includegraphics[width=.95\linewidth]{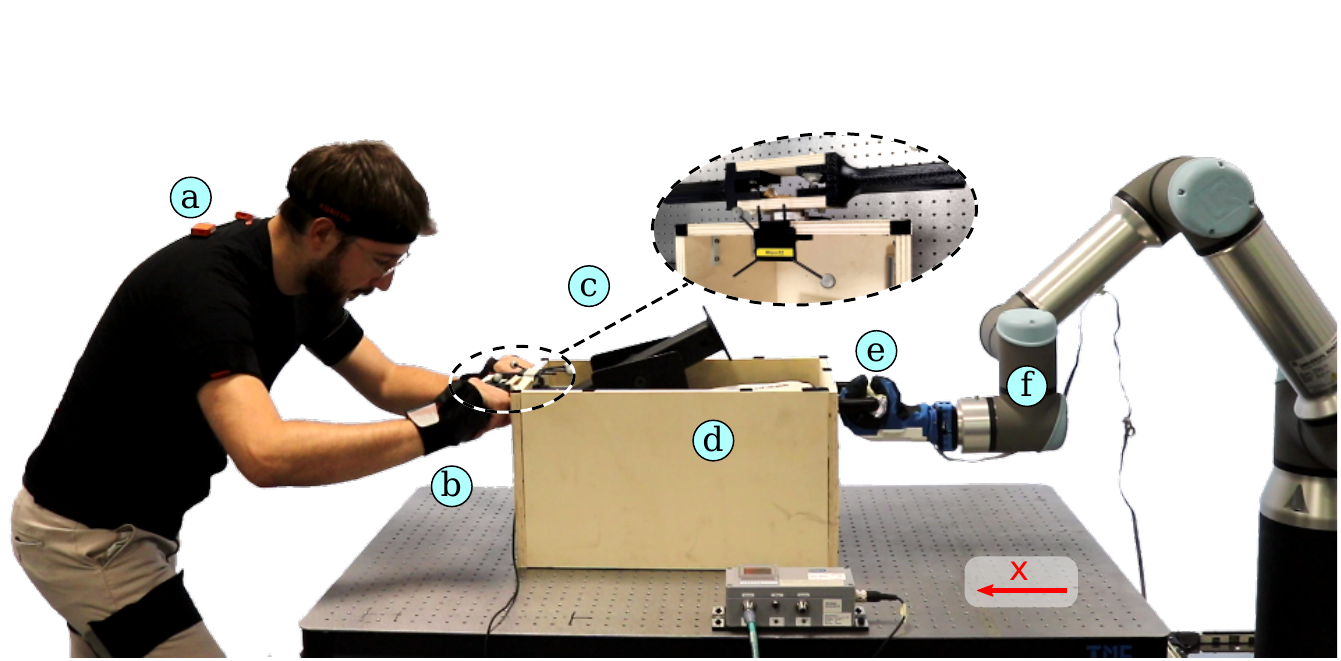}
    \caption{a) The user wears the Xsens markers for skeleton tracking; b) An F/T sensor is attached between the human-side handle and the box; c) An Optitrack marker is attached to the object for tracking its pose $\boldsymbol{x}_b$; d) A rigid wooden box filled with heavy items; e) The robot grasps the other handle using an anthropomorphic robotic hand; f) A 6-DoF position-controlled robot arm with a wrist F/T sensor. We also indicate the $x$-axis of the robot's reference frame.}
    \label{fig:exp-setup}
    \vspace{-0.2cm}
\end{figure}
\begin{table}[t]
\caption{Description of the experiments}
\centering
\begin{tabular}{lcccc}
\hline
\textbf{Exp.} & Participant & $m_{box}$ [kg] & Task time [s] & $f_{com}$ [N] \\ \hline
1        &  1         & 27.7                   & 6            & 65 \\
2        &  1         & 27.7                   & 10           & 65 \\
3        &  1         & 36.0                   & 6            & 80 \\
4        &  1         & 36.0                   & 10           & 80 \\
5        &  2         & 36.0                   & 6            & 80 \\
6        &  2         & 36.0                   & 10           & 80 \\\hline
\end{tabular} \label{tab:exp-cases}
\vspace{-0.6cm}
\end{table}

\subsection{Assistive Controller}\label{sec:assistive-controller}

We use the predicted intention $\boldsymbol{i}_h$ and the compensation force $f_{com}$ information in the assistive controller to determine the desired force $\boldsymbol{f}_d$ (Fig.~\ref{fig:control-flow}). 
The robot acts simply as a follower, applying force in the same direction as the human intention suggests. When the predicted intention changes ($\boldsymbol{i}_h \in \{-1, 0, +1\}^3$ for \textit{pull}, \textit{idle}, \textit{push}) we set the target force accordingly as $\boldsymbol{f}_d = \boldsymbol{i}_h f_{com}$. However, since the values of $\boldsymbol{i}_h$ are discrete, we linearly change the force in a transition phase that lasts 1 \textit{s}. Also, note that we set the $y$ and $z$ axes of $\boldsymbol{i}_h$ to zero in this work.

We filter the intention output due to the high variance in state transitions between idle and push-pull states and noise in human pose and F/T measurements. For this purpose, we use the average value of the last 15 values of the outputs $y_j$. 
The prediction runs at 100 \textit{Hz}, thus the filter delay is 75 \textit{ms}.

The experiments use a position-controlled 6-DoF Universal Robot model UR16. The robot's end-effector has a coupled F/T sensor, and the compliant SoftHand is used as an end-effector to grasp the box's handle. The low-level force control model is presented in the following section.

\subsection{Force Controller} \label{sec:force-control}

Our position-based force controller outputs a correction on position given an error on force: 
\begin{equation}\label{eq:force-controller}
    \boldsymbol{u} = K_f(\boldsymbol{f}_{d}-\boldsymbol{f}_{r}), \quad
    \boldsymbol{f}_{d}=\boldsymbol{i}_h f_{com}.
\end{equation}

\noindent where $\boldsymbol{f}_d$ and $\boldsymbol{f}_{r}$ are the desired force and the force measured by the robot, $K_f$ is a proportional force gain and $\boldsymbol{u}$ is the control action as a correction in position. We set $\boldsymbol{f}_d$ as the intention detection index $\boldsymbol{i}_h$ multiplied by the magnitude of an assistive force $f_{com}$ to compensate for friction force. 
We use the same $f_{com}$ value in both static and motion cases, which can lead to more help more than the friction compensation in dynamic cases.


\section{Experiments}

We run real-robot human interaction experiments to validate our system as a whole. Our experiment setup consists of a stable plane (metal table), a heavy object (a wooden box filled with objects), a human user, and a robotic assistant (6-DoF robot arm), as depicted in Fig.~\ref{fig:exp-setup}. Our sensors include the Xsens markers for skeleton tracking, an F/T sensor on the human-side handle, another F/T sensor on the robot's wrist, and an Optitrack marker for object tracking. Note that our approach utilizes only the robot-side F/T sensor and the Xsens markers (Fig.~\ref{fig:control-flow}), meaning that these two are sufficient for the system to function.
The rest are used only for evaluation and training purposes.

We evaluate our system under $6$ conditions as listed in Table~\ref{tab:exp-cases}, including different pushing speeds, different weights of the box ($m_{box}$), and different human participants (male and female).
We adjusted the speed by changing the push duration while keeping the pushing distance constant ($30$~\textit{cm}) as marked on the table. The participant is given voice feedback every second using a speaking timer app.
We adjusted the weight by adding arbitrary items in the box. The reported weights in Table~\ref{tab:exp-cases} include the box and the contents. 
Two volunteers participated in the experiments (Participant 1: Female, 155 cm, 45 kg; Participant 2: Male, 186 cm, 88 kg). A single deep learning model is trained and used for both. For this purpose, we collected our own dataset, as detailed in the next section.

We interactively determined the values of desired assistive force ${f}_{com}$ as the force compensating for the friction force without moving the object significantly (Table~\ref{tab:exp-cases}). 
For this purpose, the $f_{com}$ was iteratively changed so that the robot could almost move the object by itself. 
We name this step as \textit{object exploration} in Fig.~\ref{fig:control-flow}. It is done once before the interaction and its result is used by the controller (thus, connected by a dashed line).
In an industrial application, this step can be replaced by an automated friction estimation by adaptation methods such as \cite{Boegli2014frictionGradient, Ruderman2015frictionObserver}. 

In each experiment, the participant is asked to pull and push the box 5 times on a linear path for a given duration between the drawn marks on the table. The robot is excluded in the \textit{dry} approach; in the \textit{assisted} approach, the robot holds the box and applies force along the predicted intention. 
The experiment video is available online at \href{https://youtu.be/qy7l_wGOyzo}{https://youtu.be/qy7l\_wGOyzo}.

\begin{figure}[t]
    \centering
    \includegraphics[width=.96\linewidth]{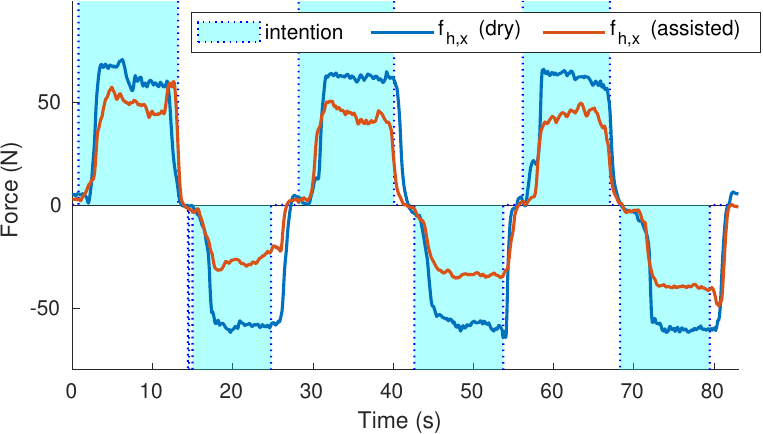}
    \caption{Comparison of the force measurements along the push direction with reference to the intention prediction during \textit{exp~2}. Cyan areas indicate the real-time pull and push prediction. The time series are shifted to align horizontally. 
    }\label{fig:intention-force-plot}
    \vspace{-0.4cm}
\end{figure}

\subsection{Dataset}\label{sec:dataset}

For the data collection, we created the same setup as shown in Fig.~\ref{fig:exp-setup}, excluding the robot. 
The participant wore the Xsens suit, the object was tracked using the Optitrack marker. 
We use Xsens as the human pose measurement system, composed of 17 IMUs placed on specific human body parts. We logged the tracked box pose $\mathbf{x}_b$ in addition to $\mathbf{X}_h$, $\boldsymbol{f}_{h}$, $\boldsymbol{f}_{r}$ and $\boldsymbol{i}_h$ in our dataset. We numerically calculated the box velocity $\mathbf{\dot{x}}_b$, and used it to determine the intention labels.

The participant was asked to pull and push the object between predetermined points involving straight, left-ward, and right-ward directions. Straight motions comprise the majority of the data. 
We included different speeds of action by counting the motion time aloud.
We also recorded idle actions like waving and exercising as negative examples.

The data includes 250$\sim$ individual pull/push actions in 28 recordings, divided roughly equally between two participants. 22 recordings are used for training, and 6 are used for validation. Unlike the original DGNN paper \cite{Shi2019dgnn} that assigns a single action label for each recording, we assign a label per each time frame (100 \textit{FPS}). We used the object velocity to label the data. Motion direction in $x$-axis indicates \textit{push} or \textit{pull} while not moving indicates \textit{idle}. The training set contains 98\textit{k}$\sim$ frames in total. The labels are distributed as $64.5\%$ \textit{idle}, $18.0\%$ \textit{pull}, and $17.5\%$ \textit{push}.  

\subsection{Results}\label{sec:results}

First, we trained a DGNN model for the intention prediction task. The final model obtains a $93.4\%$ accuracy and $94.1\%$ \textit{balanced accuracy} \cite{brodersen2010balanced} on the validation set. Although we observed higher accuracy with different parameters, we selected the one with the highest \textit{balanced accuracy} given the unbalanced nature of our dataset. The model was optimized manually to select the best learning rate ($0.02$), weight decay ($0.005$), and architecture described in Sec.~\ref{sec:intention-prediction}. Only the spatial data stream is used in this work. The motion data (time difference of joint properties) produced a lower accuracy, possibly due to noisy differential values.

\begin{figure}[t]
    \centering
    \includegraphics[width=.97\linewidth]{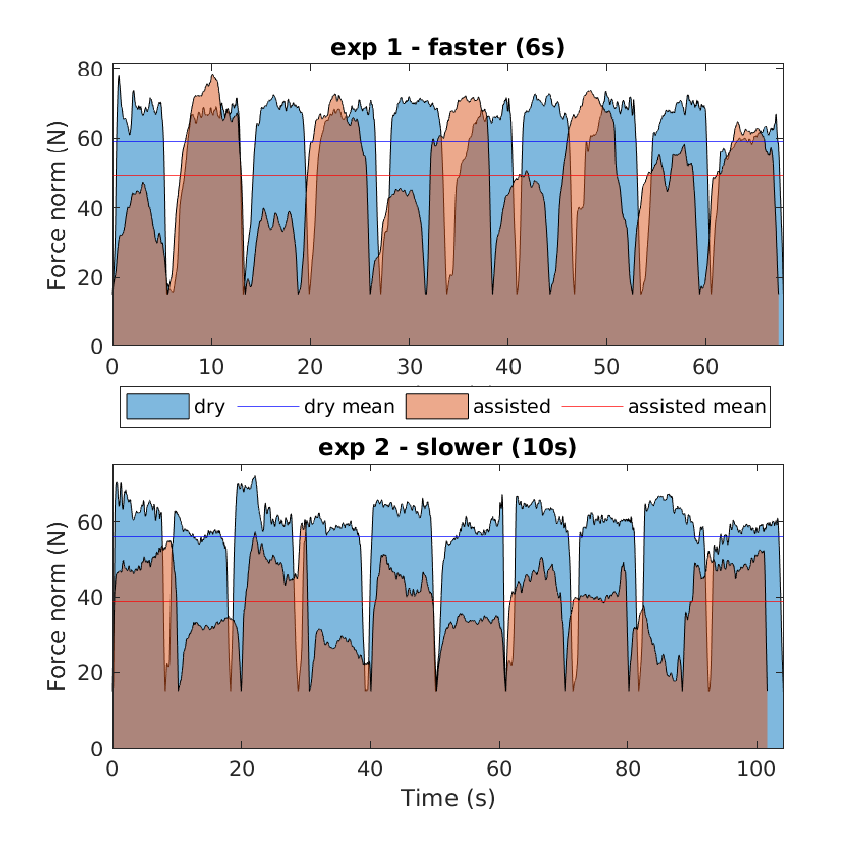}
    \vspace{-0.3cm}
    \caption{Comparison of the L2-norms of 3-D force measurements during \textit{dry} and \textit{assisted} approaches of experiments \textit{exp~1} and \textit{exp~2}. We removed the frames with less than 15~N force to filter out the idle time and consider only the significant interactions with the box. We also plot the mean of each series as a horizontal line.}\label{fig:trimmed-norms-plot}
    \vspace{-0.1cm}
\end{figure}

\begin{figure}[t]
    \centering
    \includegraphics[width=.99\linewidth]{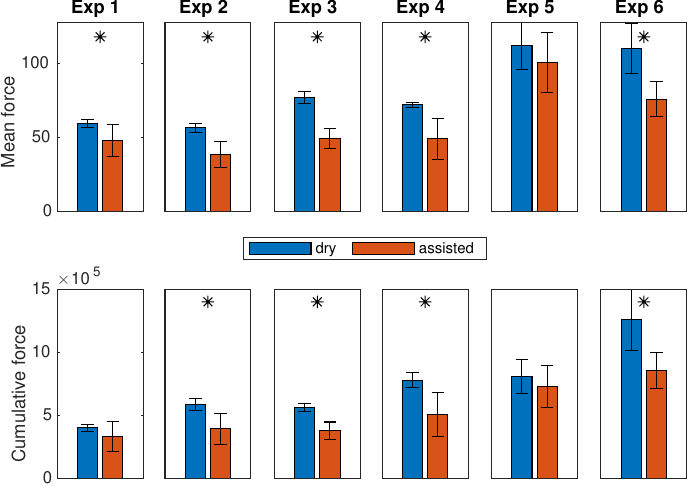}
    \caption{Mean and cumulative force exerted by the human in 10 repeated pull or push actions of each experiment case. Please see Table~\ref{tab:exp-cases} for the experiment cases. $\ast$ indicates statistical significance according to two-sample t-test with unequal variances ($p<0.01$).}\label{fig:force-stats}
    \vspace{-0.4cm}
\end{figure}

After obtaining a working prediction model, we conducted the real-world pHRI experiments as listed in Table~\ref{tab:exp-cases}. Fig.~\ref{fig:intention-force-plot} shows the force sensor data ($x$-axis) with and without assistance. As also seen on the figure, the intention is predicted before the user applies significant force and is maintained stably until the action is finished. 
On average, the intention was detected $1.54\pm 0.54$ \textit{s} ahead for participant 1 and $0.02\pm 0.33$ \textit{s} ahead for participant 2.

To quantify the total benefit of the assistance, we take the $L2$-norm of the human-side force and remove the values below 15~\textit{N} to discard the idle time and insignificant interactions with the box. 
Examples of the resulting plots are given for \textit{exp~1} and \textit{exp~2} (see Table~\ref{tab:exp-cases} for details) in Fig.~\ref{fig:trimmed-norms-plot}. 
We also show the mean value of the force during that experiment. Then, we separately calculate the average and cumulative force for each pull/push action. Mean force and the cumulative force (area under the curve) values are given for all experiments in Fig.~\ref{fig:force-stats}, with their standard deviation through 10 actions.

\subsection{Discussion}

The results validate the benefit of our collaborative robotic approach. 
The intention prediction module works stably and informs the assistive controller to synchronize efforts with the human, as demonstrated in Fig.~\ref{fig:intention-force-plot}. 
We see a significant decrease in the human effort for most of the experiments, as shown in Fig.~\ref{fig:force-stats}. Below, we discuss the insights we gained from the experiments and some of our system's limitations.

The learned DGNN model is run in real time and predicts the user's intention successfully most of the time. However, we observed a small number of discontinuities in intention prediction during the motion. 
Some failure cases are captured in Fig.~\ref{fig:intention-force-plot}: \textit{1)} at $\sim$15\textit{s}, the intention prediction changes quickly before the motion starts but does not affect the control. This type of discontinuity happened in 10 out of 60 actions. 
\textit{2)} at $\sim$80\textit{s} the \textit{push} prediction stops early, thus the user applies a larger force in the last part of the motion. Similarly, the discontinuity happened a couple more times during the motion, increasing the human effort for the rest of the action. These happened only 3 out of 60 times in total. 

The system works consistently among different object weights ( i.,e., different frictional forces) and sliding speeds, given the results of Fig.~\ref{fig:force-stats}. We share our detailed insights regarding the weight ($m$), experiment pace, and participant variations among the cases below (see Table~\ref{tab:exp-cases}). 

\begin{figure}[t]
    \centering
    \includegraphics[width=.97\linewidth]{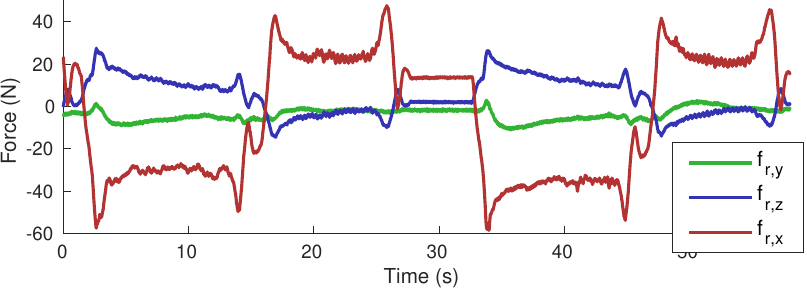}
    \caption{Robot-side force $\boldsymbol{f}_{r}$ measurements from \textit{exp 4} during two pull and two push motions. The force direction is inverse of the human-side force that is shown in Fig.~\ref{fig:intention-force-plot}.}\label{fig:robot-force-plot}
    \vspace{-0.5cm}
\end{figure}

Decreasing the velocity does not greatly impact the mean forces without assistance. Looking at the top row on Fig. \ref{fig:force-stats}, experiment pairs \textit{1-2}, \textit{3-4}, and \textit{5-6} have similar mean force without any assistance, despite the changed motion speed. However, we can see a slight increase in the assisted efforts of \textit{exp 1} and \textit{exp 5} \hide{($T{=}6$\textit{s}) }in comparison to their slower counterparts \textit{exp 2} and \textit{exp 6}\hide{ ($T{=}10$\textit{s})}. This is due to the delays in intention prediction and the low force control gain. Fig.~\ref{fig:robot-force-plot} shows the lag due to low force gain: When the motion starts the measured force reaches the desired one, however, when the human starts applying force in the same direction, the load decreases and a discrepancy occurs. The speed significantly impacts the cumulative force since slower movements take more time to execute. 

The object weight clearly impacts the base effort without assistance. Comparing \textit{exp 1-2}  (27.7 \textit{kg}) and \textit{exp 3-4} (36 \textit{kg}) for participant 1 in Fig. \ref{fig:force-stats}, we see a clear increase in the mean force for the \textit{dry} case. Meanwhile, the \textit{assisted} effort has no significant difference. This indicates that the method's contribution to context-aware intention detection tends to benefit more as the manipulated mass increases. 

We can also see distinct efforts from different participants. The participant 2 (\textit{exp 5-6}) spends more effort than the participant 1 (\textit{exp 3-4}) comparing the same mass/speed cases in Fig.~\ref{fig:force-stats}.
The distinction happens for both \textit{dry} and \textit{assisted} approaches, indicating that it is due to the participant more than the system.
The extra force can be attributed to redundant forces in $y$ and $z$ axes, which do not serve the task. 
The fact that the participants are physically diverse and that participant 1 is more familiar with the system may have influenced this outcome. 
The reduced benefit of the assistance in \textit{exp 5} can also be related to lower familiarity.

Apart from the overall performance, there are notable variations between individual pulls and pushes, as seen in Fig.~\ref{fig:trimmed-norms-plot}. 
The participants felt an asymmetry between pulling and pushing during the experiments, which makes it harder to push than pull or vice versa. In Fig.~\ref{fig:trimmed-norms-plot}, the pushes are harder in \textit{exp 1} while pulls are harder in \textit{exp 2} for the user.
This asymmetry is due to the sensor bias removal after each action. 
Because the compliant hand acts like a spring between the robot arm and the object (storing energy), it causes a bias in robot side force sensor and leads to oscillation. 
For this reason, we implemented a bias removal mechanism in the controller that every time the robot comes to a stop due to \textit{idle} prediction, the current reading of the sensor is reset to be zero reference. 
However, this lead the effective desired force to be biased and less helpful towards one side.
Fig.~\ref{fig:robot-force-plot} shows the bias in the robot-side measured force, clearly seen for $f_{r,z}$ at 30 \textit{s}.

Another discussion point is the dataset's limitation. The dataset affects the model's performance and generalizability. For this study, a user-tailored dataset was obtained to prove the proposed idea, but scalability depends on generalizable models regarding context-aware intention detection to accommodate more users.  
This may require personalizing different individuals' pulling and pushing behavior through parametrization or fine-tuning the learning model.
A larger and more diverse dataset could also improve the robustness of the predictions.

\section{Conclusions}

In this work, we presented a proof-of-concept solution for the challenging problem of indirect pHRI where human intention cannot be conveyed to the robot directly through force.
We designed and analyzed the novel problem of collaboratively transporting heavy objects which is interesting both for its technical challenges and its practical use in industrial and domestic settings.
Our context-aware assistive robotic system effectively reduces the human effort in the real-world experiments which presents a valuable benchmark solution. This is the only existing solution to this problem as we know, since the common pHRI baseline admittance control is not applicable here.

In this work, we focused on proving the concept of our method in a limited setting. The system as a whole is shown to be beneficial. However, each component of the system can be further validated and improved. 
The authors plan to conduct further studies on multidimensional intention prediction and other learning approaches. RGB-based skeleton tracking has obvious advantages, but the effects of occlusion and update rates on performance should also be studied. 
The development of autonomous friction estimation routines is another challenging and important problem on its own. 
In addition to different methodologies, future works also require analyzing generalization to new subjects and the effect of familiarity with the system. 





\addtolength{\textheight}{-3cm}

\bibliographystyle{IEEEtran}
\bibliography{biblio}

\begin{thebibliography}{10}
\providecommand{\url}[1]{#1}
\csname url@samestyle\endcsname
\providecommand{\newblock}{\relax}
\providecommand{\bibinfo}[2]{#2}
\providecommand{\BIBentrySTDinterwordspacing}{\spaceskip=0pt\relax}
\providecommand{\BIBentryALTinterwordstretchfactor}{4}
\providecommand{\BIBentryALTinterwordspacing}{\spaceskip=\fontdimen2\font plus
\BIBentryALTinterwordstretchfactor\fontdimen3\font minus \fontdimen4\font\relax}
\providecommand{\BIBforeignlanguage}[2]{{%
\expandafter\ifx\csname l@#1\endcsname\relax
\typeout{** WARNING: IEEEtran.bst: No hyphenation pattern has been}%
\typeout{** loaded for the language `#1'. Using the pattern for}%
\typeout{** the default language instead.}%
\else
\language=\csname l@#1\endcsname
\fi
#2}}
\providecommand{\BIBdecl}{\relax}
\BIBdecl

\bibitem{Stergiou2019human-human-interaction}
A.~Stergiou and R.~Poppe, ``Analyzing human–human interactions: A survey,'' \emph{Computer Vision and Image Understanding}, vol. 188, p. 102799, 11 2019.

\bibitem{Liu2021coexistence}
H.~Liu and L.~Wang, ``Collision-free human-robot collaboration based on context awareness,'' \emph{Robotics and Computer-Integrated Manufacturing}, vol.~67, p. 101997, 2 2021.

\bibitem{ajoudani2018progress}
A.~Ajoudani, A.~M. Zanchettin, S.~Ivaldi, A.~Albu-Sch{\"a}ffer, K.~Kosuge, and O.~Khatib, ``Progress and prospects of the human-robot collaboration,'' \emph{Autonomous Robots}, vol.~42, no.~5, pp. 957--975, 2018.

\bibitem{humanBodyMechanics2017}
A.~Argubi-Wollesen, B.~Wollesen, M.~Leitner, and K.~Mattes, ``Human body mechanics of pushing and pulling: Analyzing the factors of task-related strain on the musculoskeletal system,'' \emph{Safety and Health at Work}, vol.~8, pp. 11--18, 3 2017.

\bibitem{Rajendran2021ergonomicsLifting}
M.~Rajendran, A.~Sajeev, R.~Shanmugavel, and T.~Rajpradeesh, ``Ergonomic evaluation of workers during manual material handling,'' \emph{Materials Today: Proceedings}, vol.~46, pp. 7770--7776, 2021.

\bibitem{Agravante2019collabCarry}
D.~J. Agravante, A.~Cherubini, A.~Sherikov, P.-B. Wieber, and A.~Kheddar, ``Human-humanoid collaborative carrying,'' \emph{IEEE Transactions on Robotics}, vol.~35, pp. 833--846, 8 2019.

\bibitem{Ajoudani2017choosingPoses}
A.~Ajoudani, N.~G. Tsagarakis, and A.~Bicchi, ``Choosing poses for force and stiffness control,'' \emph{IEEE Transactions on Robotics}, vol.~33, pp. 1483--1490, 12 2017.

\bibitem{Shi2019dgnn}
L.~Shi, Y.~Zhang, J.~Cheng, and H.~Lu, ``Skeleton-based action recognition with directed graph neural networks,'' \emph{Proceedings of the IEEE Computer Society Conference on Computer Vision and Pattern Recognition}, vol. 2019-June, pp. 7904--7913, 2019.

\bibitem{Losey2018pHRIsurvey}
D.~P. Losey, C.~G. McDonald, E.~Battaglia, and M.~K. O'Malley, ``A review of intent detection, arbitration, and communication aspects of shared control for physical human–robot interaction,'' \emph{Applied Mechanics Reviews}, vol.~70, 1 2018.

\bibitem{Wang2024vision}
X.~Wang, D.~Veeramani, F.~Dai, and Z.~Zhu, ``Context‐aware hand gesture interaction for human–robot collaboration in construction,'' \emph{Computer-Aided Civil and Infrastructure Engineering}, 4 2024.

\bibitem{Nikolakis2018coexistence}
N.~Nikolakis, K.~Sipsas, and S.~Makris, ``A cyber-physical context-aware system for coordinating human-robot collaboration,'' \emph{Procedia CIRP}, vol.~72, pp. 27--32, 2018.

\bibitem{Quintas2019cognitive}
J.~Quintas, G.~S. Martins, L.~Santos, P.~Menezes, and J.~Dias, ``Toward a context-aware human–robot interaction framework based on cognitive development,'' \emph{IEEE Transactions on Systems, Man, and Cybernetics: Systems}, vol.~49, pp. 227--237, 1 2019.

\bibitem{drolet2023learning}
M.~Drolet, J.~Campbell, and H.~B. Amor, ``Learning and blending robot hugging behaviors in time and space,'' in \emph{2023 IEEE International Conference on Robotics and Automation (ICRA)}.\hskip 1em plus 0.5em minus 0.4em\relax IEEE, 2023, pp. 12\,071--12\,077.

\bibitem{Wong2023vision-and}
C.~Y. Wong, L.~Vergez, and W.~Suleiman, ``Vision-and tactile-based continuous multimodal intention and attention recognition for safer physical human–robot interaction,'' \emph{IEEE Transactions on Automation Science and Engineering}, vol.~21, pp. 3205--3215, 2023.

\bibitem{Vaz2020humanoid1}
J.~C. Vaz and P.~Y. Oh, ``Expanding humanoid's material-handling capabilities using capture point walking,'' \emph{Proceedings of the American Control Conference}, vol. 2020-July, pp. 2082--2087, 7 2020.

\bibitem{Polverini2020humanoid2}
M.~P. Polverini, A.~Laurenzi, E.~M. Hoffman, F.~Ruscelli, and N.~G. Tsagarakis, ``Multi-contact heavy object pushing with a centaur-type humanoid robot: Planning and control for a real demonstrator,'' \emph{IEEE Robotics and Automation Letters}, vol.~5, pp. 859--866, 4 2020.

\bibitem{Murooka2015humanoid3}
M.~Murooka, S.~Nozawa, Y.~Kakiuchi, K.~Okada, and M.~Inaba, ``Whole-body pushing manipulation with contact posture planning of large and heavy object for humanoid robot,'' \emph{Proceedings - IEEE International Conference on Robotics and Automation}, vol. 2015-June, pp. 5682--5689, 5 2015.

\bibitem{Saeedvand2021humanoid4}
S.~Saeedvand, H.~Mandala, and J.~Baltes, ``Hierarchical deep reinforcement learning to drag heavy objects by adult-sized humanoid robot,'' \emph{Applied Soft Computing}, vol. 110, p. 107601, 10 2021.

\bibitem{Kolhe2010wheeled1}
P.~Kolhe, N.~Dantam, and M.~Stilman, ``Dynamic pushing strategies for dynamically stable mobile manipulators,'' \emph{Proceedings - IEEE International Conference on Robotics and Automation}, pp. 3745--3750, 2010.

\bibitem{Ozdamar2024MobilePushing}
I.~Ozdamar, D.~Sirintuna, R.~Arbaud, and A.~Ajoudani, ``Pushing in the dark: A reactive pushing strategy for mobile robots using tactile feedback,'' \emph{IEEE Robotics and Automation Letters}, vol.~9, no.~8, pp. 6824--6831, 2024.

\bibitem{Bertoncelli2020wheeled3}
F.~Bertoncelli, F.~Ruggiero, and L.~Sabattini, ``Linear time-varying mpc for nonprehensile object manipulation with a nonholonomic mobile robot,'' \emph{Proceedings - IEEE International Conference on Robotics and Automation}, pp. 11\,032--11\,038, 2020.

\bibitem{Zhou2016manipulator1}
J.~Zhou, R.~Paolini, J.~A. Bagnell, and M.~T. Mason, ``A convex polynomial force-motion model for planar sliding: Identification and application,'' \emph{Proceedings - IEEE International Conference on Robotics and Automation}, vol. 2016-June, pp. 372--377, 2016.

\bibitem{Stuber2020survey}
J.~Stüber, C.~Zito, and R.~Stolkin, ``Let's push things forward: A survey on robot pushing,'' \emph{Frontiers in Robotics and AI}, vol.~7, 2 2020.

\bibitem{Wang2023manipulator2}
M.~Wang, A.~{\"O}. {\"O}nol, P.~Long, and T.~Pad{\i}r, ``Contact-implicit planning and control for non-prehensile manipulation using state-triggered constraints,'' in \emph{Robotics Research}, A.~Billard, T.~Asfour, and O.~Khatib, Eds.\hskip 1em plus 0.5em minus 0.4em\relax Cham: Springer Nature Switzerland, 2023, pp. 189--204.

\bibitem{Ng2024PolicyObstacle}
E.~Ng, Z.~Liu, and M.~Kennedy, ``Diffusion co-policy for synergistic human-robot collaborative tasks,'' \emph{IEEE Robotics and Automation Letters}, vol.~9, pp. 215--222, 1 2024.

\bibitem{Sirintuna2024Obstacle-awareCollabTransportation}
\BIBentryALTinterwordspacing
D.~Sirintuna, T.~Kastritsi, I.~Ozdamar, J.~M. Gandarias, and A.~Ajoudani, ``Enhancing human–robot collaborative transportation through obstacle-aware vibrotactile warning and virtual fixtures,'' \emph{Robotics and Autonomous Systems}, vol. 178, p. 104725, 2024. [Online]. Available: \url{https://www.sciencedirect.com/science/article/pii/S0921889024001088}
\BIBentrySTDinterwordspacing

\bibitem{Madan2015hapticsCarry}
C.~E. Madan, A.~Kucukyilmaz, T.~M. Sezgin, and C.~Basdogan, ``Recognition of haptic interaction patterns in dyadic joint object manipulation,'' \emph{IEEE Transactions on Haptics}, vol.~8, pp. 54--66, 1 2015.

\bibitem{Sirintuna2023MultiAgent}
D.~Sirintuna, I.~Ozdamar, and A.~Ajoudani, ``Carrying the uncarriable: a deformation-agnostic and human-cooperative framework for unwieldy objects using multiple robots,'' \emph{2023 IEEE International Conference on Robotics and Automation (ICRA)}, pp. 7497--7503, 5 2023.

\bibitem{Sirintuna2024DeformationCarry}
D.~Sirintuna, A.~Giammarino, and A.~Ajoudani, ``An object deformation-agnostic framework for human–robot collaborative transportation,'' \emph{IEEE Transactions on Automation Science and Engineering}, vol.~21, pp. 1986--1999, 4 2024.

\bibitem{Bona2005frictionOverview}
B.~Bona and M.~Indri, ``Friction compensation in robotics: an overview,'' \emph{Proceedings of the 44th IEEE Conference on Decision and Control}, pp. 4360--4367, 2005.

\bibitem{sun2022human}
Z.~Sun, Q.~Ke, H.~Rahmani, M.~Bennamoun, G.~Wang, and J.~Liu, ``Human action recognition from various data modalities: A review,'' \emph{IEEE transactions on pattern analysis and machine intelligence}, vol.~45, no.~3, pp. 3200--3225, 2022.

\bibitem{Boegli2014frictionGradient}
\BIBentryALTinterwordspacing
M.~Boegli, T.~D. Laet, J.~D. Schutter, and J.~Swevers, ``A smoothed gms friction model suited for gradient-based friction state and parameter estimation,'' \emph{IEEE/ASME Transactions on Mechatronics}, vol.~19, pp. 1593--1602, 10 2014. [Online]. Available: \url{http://ieeexplore.ieee.org/document/6675071/}
\BIBentrySTDinterwordspacing

\bibitem{Ruderman2015frictionObserver}
M.~Ruderman and M.~Iwasaki, ``Observer of nonlinear friction dynamics for motion control,'' \emph{IEEE Transactions on Industrial Electronics}, vol.~62, pp. 5941--5949, 9 2015.

\bibitem{brodersen2010balanced}
K.~H. Brodersen, C.~S. Ong, K.~E. Stephan, and J.~M. Buhmann, ``The balanced accuracy and its posterior distribution,'' in \emph{2010 20th international conference on pattern recognition}.\hskip 1em plus 0.5em minus 0.4em\relax IEEE, 2010, pp. 3121--3124.

\end{thebibliography}

\end{document}